\documentclass[10pt,twocolumn,letterpaper]{article}

\usepackage{wacv}
\usepackage{times}
\usepackage{epsfig}
\usepackage{graphicx}
\usepackage{amsmath}
\usepackage{amssymb}
\usepackage{booktabs}
\usepackage{url}            
\usepackage{booktabs}       
\usepackage{amsfonts}       
\usepackage{nicefrac}       
\usepackage{microtype}      
\usepackage{color}
\usepackage{wrapfig}
\usepackage{graphicx}
\usepackage{epstopdf}
\usepackage{amsmath,amssymb} 
\usepackage{cuted}
\usepackage{bbm}
\usepackage{multirow}
\usepackage{makecell}
\usepackage{multirow}
\usepackage{soul}
\usepackage{arydshln}
\usepackage{pifont}
\usepackage[accsupp]{axessibility}  

\def\wacvPaperID{1532} 

\wacvapplicationstrack 

\wacvfinalcopy 

\ifwacvfinal
\usepackage[breaklinks=true,bookmarks=false]{hyperref}
\else
\usepackage[pagebackref=true,breaklinks=true,colorlinks,bookmarks=false]{hyperref}
\fi

\begin{document}

\newcommand{\xmark}{\ding{55}}%
\newcommand{\cmark}{\ding{51}}%

\title{EfficientPhys: Enabling Simple, Fast and Accurate \\ Camera-Based Cardiac Measurement} 

\author{Xin Liu\textsuperscript{1}, Brian Hill\textsuperscript{2}, Ziheng Jiang\textsuperscript{1}, Shwetak Patel\textsuperscript{1}, Daniel McDuff\textsuperscript{3}\\
Paul G. Allen School of Computer Science \& Engineering, University of Washington, Seattle, WA, USA\textsuperscript{1} \\
Department of Computer Science, University of California, Los Angeles, CA, USA\textsuperscript{2} \\
Microsoft Research, Redmond, WA, USA\textsuperscript{3} \\ 
{\tt\small \{xliu0, shwetak, ziheng\}@cs.washington.edu} \\
{\tt\small brian.l.hill@cs.ucla.edu, damcduff@microsoft.com}
}

\maketitle
\thispagestyle{empty}
\begin{abstract}

Camera-based physiological measurement is a growing field with neural models providing state-of-the-art performance. Prior research has explored various ``end-to-end'' architectures; however these methods still require several preprocessing steps and are not able to run directly on mobile and edge devices. The operations are often non-trivial to implement, making replication and deployment difficult and can even have a higher computational budget than the ``core'' network itself. In this paper, we propose two novel and efficient neural models for camera-based physiological measurement called EfficientPhys that remove the need for face detection, segmentation, normalization, color space transformation or any other preprocessing steps. Using an input of raw video frames, our models achieve strong accuracy on three public datasets. We show that this is the case whether using a transformer or convolutional backbone. We further evaluate the latency of the proposed networks and show that our most lightweight network also achieves a 33\% improvement in efficiency. 
\end{abstract}

\section{Introduction}

Camera-based physiological measurement is a non-contact approach for capturing cardiac signals via light reflected from the body~\cite{mcduff2021camera}. The most common such signal is the blood volume pulse (BVP) measured via the photoplethysmogram (PPG). From this, heart rate~\cite{takano2007heart,verkruysse2008remote}, respiration rate~\cite{poh2010advancements} and pulse transit times~\cite{shao2014noncontact} can be derived. Furthermore, there is promising evidence that the PPG signals can be used to measure signs of arterial disease~\cite{takazawa_assessment_1998}. Neural models are the current state-of-the-art for camera PPG measurement~\cite{chen2018deepphys,liu2020multi,liu2021video}. These networks can learn strong feature representations and effectively disentangle the subtle changes in pixels due to underlying physiological processes from those due to body motions, lighting changes and other sources of ``noise''.

While prior research has framed architectures as ``end-to-end'' methods, those that achieve state-of-the-art performance actually require several preprocessing steps before data is used as input to the network.  
For example, \cite{chen2018deepphys} and ~\cite{liu2020multi} use hand-crafted normalized difference frames and normalized appearance frames as input to their convolutional attention network. \cite{niu2020video} and \cite{lu2021dual} use a complex schema to create feature maps called ``MSTmaps'', their process includes facial landmark detection, extraction of several regions of interest (ROI) using these landmarks, and then averaging pixel values in both the RGB and YUV color spaces.

These preprocessing steps have several drawbacks: 1) They make assumptions about optimal normalization or representation without allowing the network to learn these features in a data-driven manner. 2) They are computationally costly and in many cases add a significant number of operations to the video processing pipeline. There are several reasons why running camera-based physiological sensing on-device is desirable: privacy preservation, the ability to use raw (i.e., uncompressed) video and data cost and bandwidth savings. Any additional computation needs to be justified by improving model accuracy. Moreover, since camera-based physiological sensing is a privacy-sensitive application, it is preferable to store the data on local devices instead of streaming both video and physiological data to the cloud. The overhead from processing is not acceptable if we aim to make the system accessible to low-end mobile devices. 3) Many of these steps are non-trivial to implement and optimize in and of themselves. This makes it harder to deploy real-time systems and to replicate the implementation on different platforms. For instance, implementing existing methods on Android, iOS, or in JavaScript requires a significant amount of effort. Some libraries, such as facial landmark detection, are not even available on every platform. Thus, the last mile engineering using the existing methods becomes especially challenging.

\begin{figure*}[t!]
  \includegraphics[width=\textwidth]{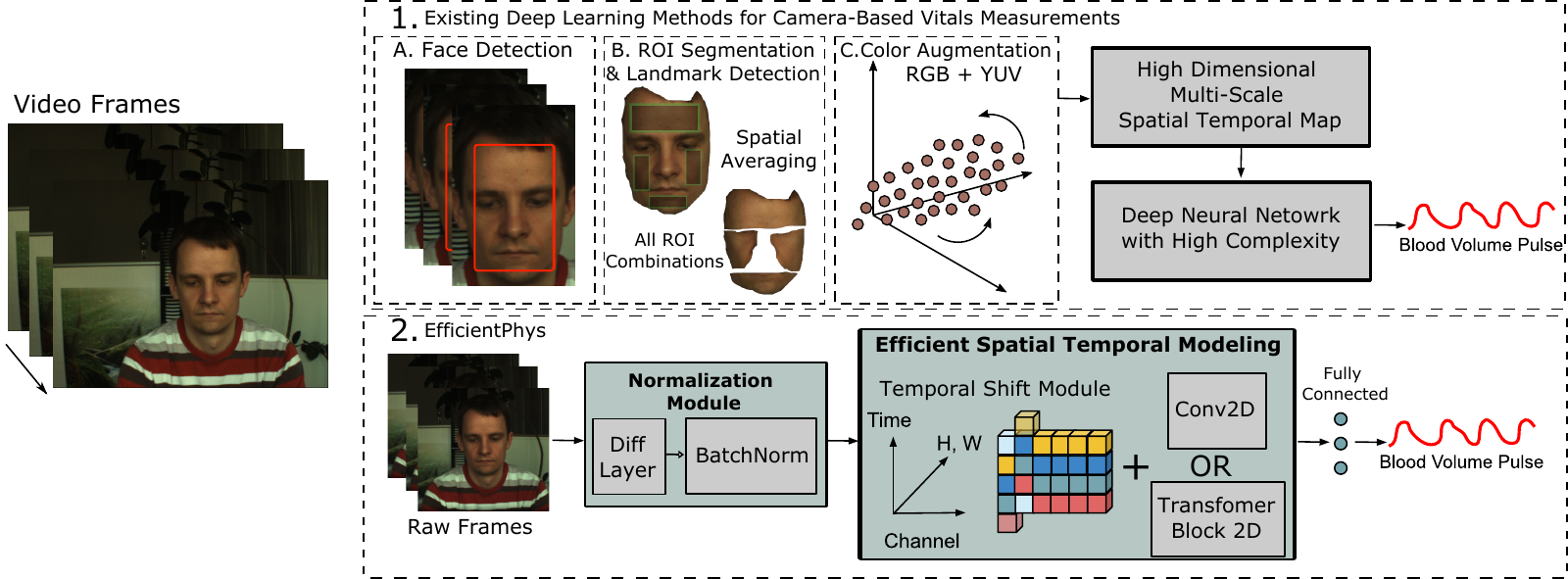}
  \caption{A high-level comparison of EfficientPhys and existing deep learning approaches for camera-based vitals measurement}
  \label{fig:preprocess_comp}
\end{figure*}

Ideally, a video-based physiological measurement method would be able to run at a high frame rate even on mobile devices, be simple to implement across different platforms, and achieve state-of-the-art performance. Addressing the aforementioned challenges would help achieve these properties. We propose a truly end-to-end network, EfficientPhys, for which the input is unprocessed video frames without requiring accurate face cropping (see Fig.~\ref{fig:preprocess_comp}). 
Due to recent advancements in visual transformers, we propose both a convolutional and visual transformer architecture and compare and contrast the performance of these two. 

In summary, our key contributions are to: 1) propose two novel one-stop neural architectures, a visual transformer and a convolutional network, which do not require any preprocessing steps, 2) evaluate the proposed methods on three popular benchmark datasets, 3) evaluate on-device latency across both state-of-the-art machine learning-based approaches as well as signal processing-based techniques. To the best of our knowledge, this is the first paper that explores the visual transformer in camera-based physiological measurement and its comparison with convolutional networks. This is also the first paper exploring a completely end-to-end on-device neural architecture for mobile devices. 
\section{Related Work}

\textbf{Camera-based Vital Measurement.} 
There is a growing community studying the use of cameras to sense physiological vitals signs \cite{wu2000photoplethysmography,verkruysse2008remote, liu2021camera}. Prior work established the fundamentals of how RGB images could be used to extract the pulse signal using signal source separation techniques (e.g., ICA) \cite{poh2010advancements}. Other methods derived these parameters from physically-based models to achieve elegant and fast demixing (e.g., Plane Orthogonal-to-Skin (POS))\cite{wang_algorithmic_2017}. By calculating a projection plane orthogonal to the skin-tone based on optical and physiological principles, the authors were able to achieve a stronger BVP signal-to-noise ratio (SNR).

Since the underlying relationship between the pulse and skin pixels is complex, deep convolutional neural networks have shown superior performance over the traditional source separation algorithms. DeepPhys \cite{chen2018deepphys} was the first paper that demonstrated that a deep neural network outperforms all the traditional signal processing approaches. Liu et al. have also proposed an on-device efficient neural architecture called MTTS-CAN for camera-based physiological sensing, which leverages a tensor-shift module and 2D-convolutional operations to perform efficient spatial-temporal modeling \cite{liu2020multi}. More recently, an adversarial learning approach, called Dual-GAN, has also been studied to learn noise-resistant mappings from video frames to pulse waveform and noise distributions \cite{lu2021dual}. With two generative-adversarial networks, they can promote each adversarial network's representation and further improve the feature disentanglement between pulse and various noise sources. 

However, DeepPhys and MTTS-CAN both require a few preprocessing steps including calculating difference frames and performing image normalization. Dual-GAN has a even more complex preprocessing module called MSTMaps proposed by \cite{niu2020video}. The MSTMaps are a set of multi-scale spatial temporal maps created by 1) cropping the facial region, 2) extracting facial landmarks, 3) performing average pooling for every color channel and every ROI combination for each frame, 4) generating ROI combinations using all the detected ROI regions and landmarks, 5) multiplying each item in all ROI combinations with six channels. The final size of the MSTMap is $(2^n - 1) \times T \times 6$ where $T$ is the number of frames and $n$ is the number of ROI regions. Such a preprocessing module not only consumes large amounts of memory but also introduces a large computational burden to the entire pipeline. Moreover, stacking all of these extra procedures makes development and deployment much more difficult.  Unlike these methods, the goal of our proposed method EfficientPhys is to create a preprocessing-free neural architecture that is simple to use and deploy, efficient on mobile devices, and accurate on settings with various types of noise.

\textbf{Visual Transformers.} Although convolutional neural networks have been widely studied and used in many computer vision applications, vision transformers were first successfully used on the task of image classification. By training on larger datasets, vision transformer (ViT) attains excellent performance and can be used in downstream fine-tuning with fewer amounts of data \cite{dosovitskiy2020image}. More recently, the state-of-the-art Swin vision transformer was proposed as a way to construct hierarchical feature maps and improve computational efficiency by using a hierarchical representation and limiting self-attention computation to non-overlapping local windows while allowing for cross-window connection \cite{liu2021swin}. However, transformer architectures have been barely studied in the field of camera-based vitals measurement. The closest work used transformers to detect remote photoplethysmography (rPPG) for attack/spoofing detection ~\cite{yu2021transrppg}. However, this paper did not evaluate the proposed vision transformer in the task of heart rate estimation using any public datasets, which is considered as the gold-standard benchmark for the field of camera-based vital measurement. More recently, Yu \textit{et al.} recently proposed Physformer \cite{yu2022physformer} which is also a visual transformer based architecture \footnote{Physformer was published after this manuscript was submitted for review}. To our best knowledge, our proposed vision transformer was
the first architecture in camera-based heart rate measurement with detailed
evaluation on multiple public datasets.

\section{Method}

\begin{figure*}[t!]
  \includegraphics[width=\textwidth]{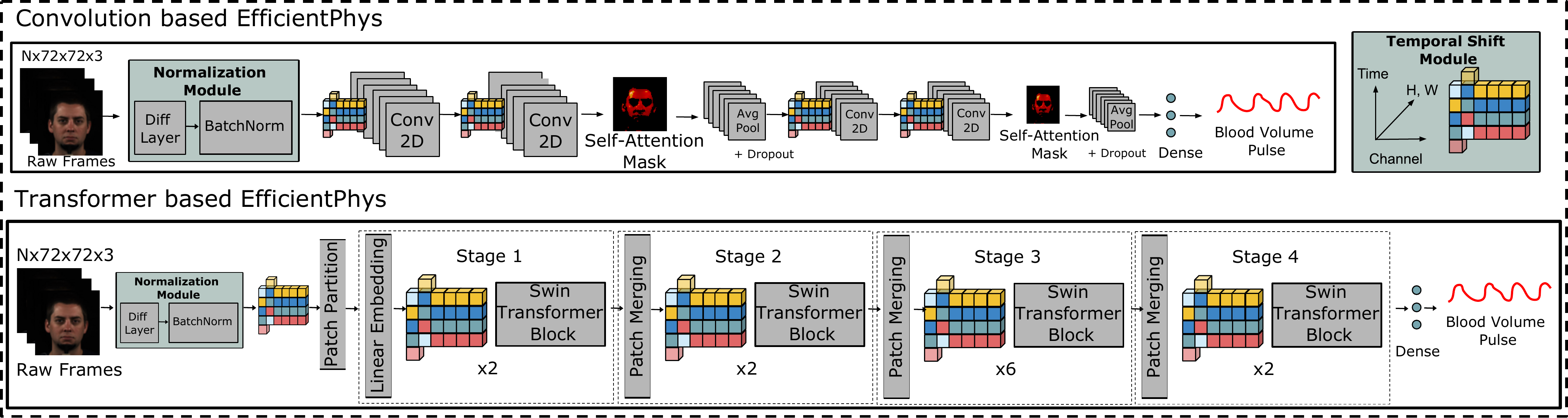}
  \caption{We present two novel architectures to enable simple, fast, and accurate camera-based vitals measurement: Convolution-based EfficientPhys and Transformer-based EfficientPhys. N is the number frames of video clip inputting to the network. }
  \label{fig:high_level_comp}
\end{figure*}

\subsection{Convolution-based EfficientPhys}

To enable simple, fast and accurate real-time on-device camera-based vitals measurement, we propose a one-stop solution architecture that takes raw video frames as the input to the network and outputs a first-derivative PPG signal. The convolution-based EfficientPhys is a one-branch network that contains a custom normalization layer, self-attention module, tensor-shift module and 2D convolution operation to perform efficient and accurate spatial-temporal modeling while making it simple to deploy. 

\textbf{Normalization Module.} Existing neural methods all require different levels of preprocessing before providing the visual representation to the network to learn the underlying relationship between skin pixels and cardiac pulse signal. For instance, The state-of-the-art networks Dual-GAN \cite{lu2021dual} and CVD \cite{niu2020video} proposed a hand-crafted spatial-temporal representations called STMaps. These preprocessed representations are generated for each video frame and includes steps of detecting 81 facial landmark points, extracting a set of region of interest (ROI) combinations ($2^n -1$ where n is the number of ROIs, n=6) using these landmarks, and averaging pixel values in both the RGB and YUV color spaces, multiplying the 63 ROI combinations with the six channels. These modules not only add significant computational burden (Table~\ref{tab:latency} shows that Dual-GAN's preprocessing module takes 275ms per frame) but also make the system more challenging to implement and deploy on real-world computing systems such as mobile devices.  

One of the goals of EfficientPhys is to remove these preprocessing modules completely and provide a one-stop solution. To achieve such simplicity and deployability, we propose a custom normalization module, which can perform motion modeling between every two consecutive RGB raw frames and normalization to reduce the lighting and motion noise. More specifically, the proposed normalization module includes a difference layer and a batchnorm layer. The difference layer (e.g., torch.diff) computes the first forward difference along the temporal axis of the raw video frames, by subtracting every two adjacent frames. Performing motion modeling between every two consecutive frames and normalization is more like a high-pass filtering and can help reduce the global noise from lighting and motion noise, while maintaining the subtle changes from PPG. To provide optical basis in our work, equation \ref{eq:1} illustrates the optical grounding of difference frame where $\pmb{D}_k(t)$ of every two consecutive frames. $I(t)$ is the luminance intensity which is modulated by the specular reflection $\pmb{v}_s(t)$ and the diffuse reflection $\pmb{v}_d(t)$ as well as the optical sensor's quantization noise $\pmb{v}_n(t)$. 

\begin{equation} \label{eq:1}
\begin{split}
    \small
	\pmb{D}_k(t)= (I(t) \cdot (\pmb{v}_s(t)+\pmb{v}_d(t)) + \pmb{v}_n(t)) - (I(t-1) \\
	\cdot (\pmb{v}_s(t-1)+\pmb{v}_d(t-1))+\pmb{v}_n(t-1))
\end{split}
\end{equation}

\begin{figure*}[t!]
  \includegraphics[width=\textwidth]{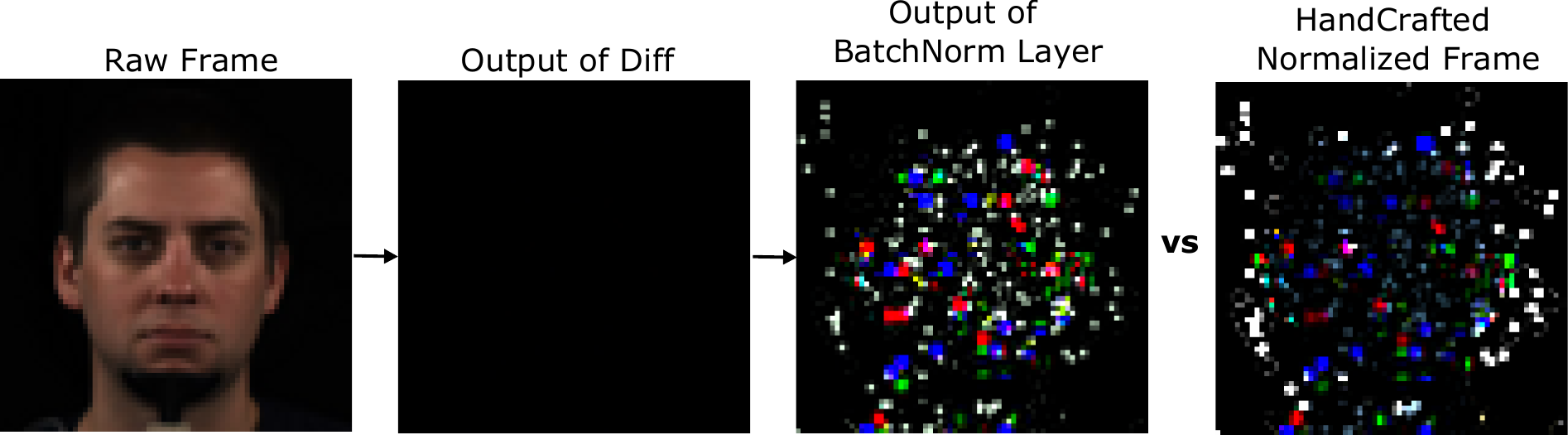}
  \caption{Outputs of diff and batchnorm layers and comparison with normalized frames generated via the hand-crafted process in prior work~\cite{chen2018deepphys}. The output from the diff layer is almost black because the difference in skin pixels of consecutive frames is very subtle. }
  \label{fig:bachnorm}
\end{figure*}

However, difference frames could be dramatically different in scale and make it hard for the network to learn meaningful feature representations, especially when the signal of interest is hidden in subtle pixel changes along the temporal axis and noise artifacts can cause significantly larger relative changes. To address this, we add a batch-normalization (batchnorm) layer following the difference layer. Adding a batchnorm layer provides two benefits: 1) it normalizes the difference frames to the same scale within the batch during training, 2) unlike fixed normalization in previous work \cite{chen2018deepphys,liu2020multi}, batchnorm provides two learnable parameters $\beta$ and $\gamma$ for scaling (to a different variance) and shifting (to a different mean) and two non-trainable parameters which are the mean $\mu$ and the standard deviation $\sigma$. Through the learning process, the batchnorm layer can learn the best parameters for amplifying the pixel changes while minimizing the noise as Equation \ref{eq:2} and Fig.\ref{fig:bachnorm} show.  Without a batchnorm layer, directly applying a difference layer means the frames appear ``black''; because the subtle changes of skin pixels in every two consecutive frames are relatively very small. On the other hand, adding a follow-up batchnorm layer will help it learn the normalization function to magnify the subtle changes of skin pixels substantially. The result is not simply a magnification of values but a normalization and magnification. Moreover, we also compare the output batchnorm layer to the hand-crafted normalized frame as shown in Fig.\ref{fig:bachnorm}. The output of batchnorm layer contains more information and qualitative analysis suggests it should be a better tool for skin segmentation after the learning process.  

\begin{equation} 
    \label{eq:2}
	\pmb{N}_k(t)= \frac{(\beta_t * \pmb{D}_k(t) + \gamma_t) - \mu_{\pmb{D}_k}}{\sigma_{\pmb{D}_k}}
\end{equation}

\textbf{Self-Attention-Shifted Network.}
To efficiently capture the rich spatial-temporal information, we propose a self-attention-shifted network (SASN). SASN is built on top of the previous state-of-the-art method for on-device spatial-temporal modeling in optical cardiac measurement - temporal-shift convolutional attention network (TS-CAN) \cite{liu2020multi}. TS-CAN has two convolutional branches, one of which takes a preprocessed difference frame representation and one of which takes a normalized appearance frame. The motion branch performs the main spatial-temporal modeling and estimation, and the appearance branch provides attention masks to guide the motion branch to better isolate the pixels of interest (e.g., skin pixels). However, we argue that the attention masks do not have to be obtained through a separate appearance branch and they can be also learned with a single branch end-to-end network. As Fig. \ref{fig:high_level_comp} illustrates, our proposed self-attention-shifted network starts with the custom normalization module discussed in the previous section and then continues with two tensor-shifted convolutional operations. After the second and fourth tensor-shifted 2D convolutional layers, we add a self-attention module to help the network minimize the negative effects introduced by temporal shifting as well as motion and lighting noise. The self-attention layers are softmax attention layers with 1D convolutions followed by a sigmoid activation function. Then, normalization is applied to remove the outlier values in the attention mask, and the final normalized attention mask is element-wise multiplied with the output from the tensor-shifted convolution. Equation \ref{eq:attention} summarizes how our self-attention mechanism works where $ts(.)$ denotes temporal shift operation, $\omega^{t}_{c}$ denotes the 2D convolutional kernel followed by the temporal shift module,  $\sigma$ is the sigmoid activation and $\omega_{a}^t$ is the 1 $\times$ 1 convolutional kernel for self attention. 

\begin{equation} \label{eq:attention}
      (\omega^{t}_{c}ts(\pmb{N}_k(t)) + b_c^t) \odot \frac{H_t W_t \cdot \sigma(\omega_{a}^t\mathbb{X}_{\alpha}^t + b_a^t)}
    {2\parallel\sigma(\omega_{a}^t\mathbb{X}_{\alpha}^t + b_a^t)\parallel_1}
\end{equation}

\subsection{Transformer-based EfficientPhys}

\textbf{Efficient Spatial-Temporal Video Transformer.} Due to the recent success of visual transformers for image and video understanding and the importance of attention mechanisms for this task \cite{yu2019remote,chen2018deepphys,qi2020deeprhythm,liu2020multi}, we also present a visual transformer version of EfficientPhys. For this task, we need a visual transformer to learn both spatial and temporal representations. Several existing video-based visual transformers are based on 3D-embedding tokens and input all the frames into 3D encoder and spatial-temporal attention modules \cite{arnab_vivit_2021,liu2021video}. However, the computational complexity makes these unfavourable for real-time efficient modeling on mobile devices. In the convolutional version we used tensor-shifted 2D convolutions which have been shown to achieve comparable performance as 3D convolutions \cite{liu2020multi}. Inspired by this, our proposed transformer-based EfficientPhys is based on a 2D visual transformer, Swin transformer \cite{liu_swin_2021}, but with additional components that we will describe below.

Since the 2D Swin transformer is only able to learn spatial
features that map raw RGB values to latent representations between a single frame and the target signal (pulse), it does not have ability to model temporal relationships beyond consecutive frames. One of the main contributions of the Swin transformer is the shifted window module which has linear computation complexity and allows cross-window connection by shifting the window partition and limiting self-attention computation to non-overlapping local windows. Inspired by the idea of shifting of spatial window partitions, we propose to add a tensor-shift module (TSM) \cite{lin2019tsm} before every Swin transformer block to facilitate information exchange across the temporal axis.
The TSM first splits the input tensor into three chunks, shifts the first chunk to the left by one place (advancing time by one frame) and shifts the second chunk to the right by one place (delaying time by one frame). All shifting operations are along temporal axis and are performed before the tensor is fed into each transformer block as shown in Fig. \ref{fig:high_level_comp}. By adding the TSM module to the Swin transformer, the new transformer architecture now has the ability to perform efficient spatial-temporal modeling and attention by combining shifting window partitions spatially and shifting frames temporally. It is worth noting that TSM does not introduce any learnable parameters, thus the proposed transformer architecture has the same number of parameters as the original Swin transformer. Finally, to enable truly end-to-end inference and learning, we also add the same normalization module proposed in the convolution EfficientPhys to this architecture. 

In summary, the transformer-based EfficientPhys is the first end-to-end transformer architecture for camera-based cardiac pulse measurement that leverages tensor-shift modules and window-partition shift modules to perform efficient spatial-temporal modeling and attention to learn the underlying physiological signal from skin pixels. 

\section{Experiments}

\textbf{Training Data.} 
To help create a robust and generalizable model for cross-dataset evaluation we use two datasets. The first is AFRL \cite{estepp_recovering_2014}, which includes 300 videos from 25 subjects (17 males and 8 females). For each video, the raw resolution is 658x492 and the sampling rate of the synchronized pulse measurement is 30Hz. The dataset includes videos with a range of head motions. Every participant was instructed to remain stationary for the first two tasks, and then to perform head motions with increasing rotational velocity in the next four tasks (turning from left to right). Along with the AFRL dataset, we also leverage a synthetic avatar video dataset introduced by \cite{mcduff_advancing_2020, mcduff2022scamps} where each synthetic video is parameterized and generated with a custom pulse signal, background, facial appearance, and motion. More specifically, the input pulse signal is used to augment skin color and the subsurface radius of skin pixels to mimic the effect of the blood volume pulse on the skin's appearance. Synthetic data such as this introduces greater diversity into the training set and has been shown to effectively help reduce disparities in performance by skin type. All the video data are resized into a resolution of 72$\times$72 with resampling using pixel area relation (\texttt{cv2.INTER\_AREA}). 

\textbf{Testing Data.} 
We use three popular benchmark datasets to evaluate the accuracy of the proposed EfficientPhys. UBFC \cite{bobbia2019unsupervised} is a dataset of 42 videos from 42 subjects, and the raw resolution of each video is 640x480 in an uncompressed 8-bit RGB format. The sampling for synchronized pulse signal is 30 Hz. All of the tasks collected in UBFC are stationary. MMSE \cite{zhang2016multimodal} is a dataset including 102 videos from 40 subjects, and the raw resolution of each video is 1040x1392. The ground-truth waveform for MMSE is blood pressure signal instead of blood-volume pulse signal, and the sampling rate is 25 Hz. It is worth noting that MMSE contains a diverse distribution of skin types in Fitzpatrick scale (II=8, III=11, IV=17, V+VI=4). PURE \cite{stricker2014non} is a dataset containing 60 videos from 10 subjects. The raw resolution of each video is 640x480, and the sampling rate of the ground-truth pulse signal is 60 Hz. PURE includes a diverse set of motion tasks such as steady, talking, slow/fast translation between head movements and the camera plane, and small/medium head rotation.

\textbf{Implementation \& Experiment Details.}
We implemented both convolution-based and transformer-based EfficientPhys in PyTorch \cite{paszke2019pytorch}. We used an AdamW optimizer to train both networks instead of Adam by introducing additional regularization to reduce the effects of over-fitting through weight decay \cite{loshchilov2017decoupled}. The learning rate we used for Convolutional model was 0.001 while the rate for transformer model was 0.0001. Based on empirical studies, we used the mean squared error (MSE) loss for training the transformer models and negative Pearson loss \cite{tsou2020siamese} for the convolutional model. We trained both models for ten epochs with a fixed random seed. We implemented TS-CAN based on the open-sourced code \cite{liu_metaphys_2021,liu2020multi} and used the Deep Physiological Sensing Toolbox \cite{liu2022deep} for the experiments on the UBFC and PURE datasets. To calculate the performance metrics, we first applied a band-pass filter to the signal with a cutoff frequency of 0.75 and 2.5 Hz (45 beats/minute to 150 beats/minute). We then followed Dual-GAN's evaluation scheme using peak detection and FFT to get estimated heart rate on each video of UBFC and PURE datasets \cite{lu_dual-gan_nodate} and MetaPhys's evaluation scheme on MMSE \cite{liu_metaphys_2021}. We conducted video-level evaluation where we calculated an averaged heart rate for each single video. We calculated three standard metrics for each video: mean absolute error (MAE), root mean squared error (RMSE) and Pearson correlation ($\rho$) in heart rate estimations and the corresponding ground-truth heart rates from the blood volume pulse collected via contact oximeter sensor.

To explore the efficiency of different architectures on mobile devices, we also conducted experiments on a quad-core Cortex-A72 Raspberry Pi 4B to evaluate the model's performance on an edge device. We performed inference 10 times to get a reliable averaged on-device inference latency for EfficientPhys and TS-CAN. Due to the lack of open-source implementation of Dual-GAN, we were only able to find the implementation of STMaps which is the preprocessing module of Dual-GAN. Thus, we only evaluated the on-device latency for the preprocessing module in Dual-GAN. We also evaluated the latency of POS, CHROM, and ICA as they are traditional signal processing methods and don't have a separate preprocessing module.

\section{Results and Discussion}

\begin{table*}[t!]
    \small
	\caption{Cross-dataset heart rate evaluation on UBFC and PURE (beats per minute).}
	\vspace{0.3cm}
	\label{tab:cross_eval_crop}
	\centering
	\small
	\setlength\tabcolsep{3pt} 
	\begin{tabular}{r|cccc|cccc}
	\toprule
	    &\multicolumn{4}{c}{\textbf{UBFC} \cite{bobbia2019unsupervised}} & \multicolumn{4}{c}{\textbf{PURE}  \cite{stricker2014non}} \\
        \textbf{Method} & MAE$\downarrow$ & MAPE$\downarrow$ &  RMSE$\downarrow$ & $\rho$ $\uparrow$ & MAE$\downarrow$ & MAPE$\downarrow$ & RMSE$\downarrow$ & $\rho$ $\uparrow$ \\ \hline  \hline 
        EfficientPhys-C & 1.14 & 1.16\% & 1.81 & 0.99 & 1.33 &  1.71\% & 5.99 & 0.97  \\
        EfficientPhys-T1 & 2.08 & 2.53\% & 4.91 & 0.96 & 1.11 & 1.30\% & 5.94 & 0.97  \\
        EfficientPhys-T2 & 3.07 & 3.41\% & 4.78 & 0.96 & 2.67 & 3.22\% & 9.08 & 0.92  \\ \hdashline
        TS-CAN\cite{liu2020multi}& 1.70 & 1.99\% & 2.72 & 0.99 & 2.23 & 2.25\% & 3.71 & 0.98  \\
        POS\cite{wang_algorithmic_2017} & 3.52 & 3.36\% & 8.38 & 0.90 & 1.68 & 1.56\% &  9.60 & 0.92 \\
        CHROM\cite{de2013robust} & 3.10 & 3.83\% & 6.84 & 0.93 & 6.23 & 10.04\% & 17.18 & 0.71  \\
        ICA\cite{poh2010advancements}& 4.39 & 4.30\% & 11.60 & 0.82 & 5.70 & 5.69\% & 18.10 & 0.70 \\
       \bottomrule 
       \end{tabular}
       \\
       \tiny
      MAE = Mean Absolute Error in HR estimation, MAPE = Mean Absolute Error Percentage in HR estimation, RMSE = Root Mean Square Error in HR estimation, $\rho$ = Pearson Correlation in HR estimation.
    \vspace{-0.1cm}
\end{table*}

\begin{table}[t!]
    \small
	\caption{Cross-dataset heart rate evaluation on MMSE (beats per minute).}
	\vspace{0.3cm}
	\label{tab:cross_eval_mmse}
	\centering
	\small
	\setlength\tabcolsep{3pt} 
	\begin{tabular}{r|cccc}
	\toprule
	    & \multicolumn{4}{c}{\textbf{MMSE}~\cite{zhang2016multimodal}} \\
        \textbf{Method} & MAE$\downarrow$ & MAPE$\downarrow$ &  RMSE$\downarrow$ & $\rho$ $\uparrow$ \\\hline  \hline 
        EfficientPhys-C & 3.48 & 4.02\% & 7.21 & 0.86  \\
        EfficientPhys-T1 & 3.04 & 3.91\% & 5.91 & 0.92  \\
        EfficientPhys-T2 & 3.51 & 3.96\% & 6.98 & 0.88 \\ \hdashline
        TS-CAN\cite{liu2020multi} & 3.04 & 3.41\% & 6.55 & 0.89  \\
        POS\cite{wang_algorithmic_2017} & 3.79 & 4.28\% & 8.47 & 0.82 \\
        CHROM\cite{de2013robust} & 3.61 & 4.50\% & 7.43 & 0.85 \\
        ICA\cite{poh2010advancements} & 7.96 & 9.20\% & 14.02 & 0.51 \\
       \bottomrule 
       \end{tabular}
       \\
       \tiny
      MAE = Mean Absolute Error in HR estimation, MAPE = Mean Absolute Error Percentage in HR estimation, RMSE = Root Mean Square Error in HR estimation, $\rho$ = Pearson Correlation in HR estimation.
    \vspace{-0.1cm}
\end{table}

\begin{table}[t!]
    \small
	\caption{Cross dataset evaluation with models trained on PURE only and tested on UBFC (beats per minute).}
	\vspace{0.3cm}
	\label{tab:cross_pure_ubfc}
	\centering
	\small
	\setlength\tabcolsep{3pt} 
	\begin{tabular}{r|cccc}
	\toprule
        \textbf{Method} & MAE$\downarrow$ &  MAPE$\downarrow$ & RMSE$\downarrow$ & $\rho$ $\uparrow$ \\ \hline  \hline 
        EfficientPhys-C & 2.13 & 2.35 \% & 3.00 & 0.99   \\
        EfficientPhys-T1 & 3.83 & 4.32\% & 5.62 & 0.87   \\
        EfficientPhys-T2 & 3.97 & 4.35\% & 5.91  & 0.94   \\
        TS-CAN\cite{liu2020multi}& 1.16 & 1.42\% & 2.78 & 0.99   \\
        Dual-GAN\cite{lu2021dual} & 0.74 & 0.73\% & 1.02 & 0.99   \\
        PulseGAN\cite{song2021pulsegan} & 2.09 & 2.23\% & 4.42 & 0.99  \\
       \bottomrule 
       \end{tabular}
       \\
       \tiny
      MAE = Mean Absolute Error in HR estimation, MAPE = Mean Absolute Error Percentage in HR estimation, RMSE = Root Mean Square Error in HR estimation, $\rho$ = Pearson Correlation in HR estimation.
    \vspace{-0.1cm}
\end{table}

\begin{figure}[t!]
\centering
\includegraphics[width=0.4\textwidth]{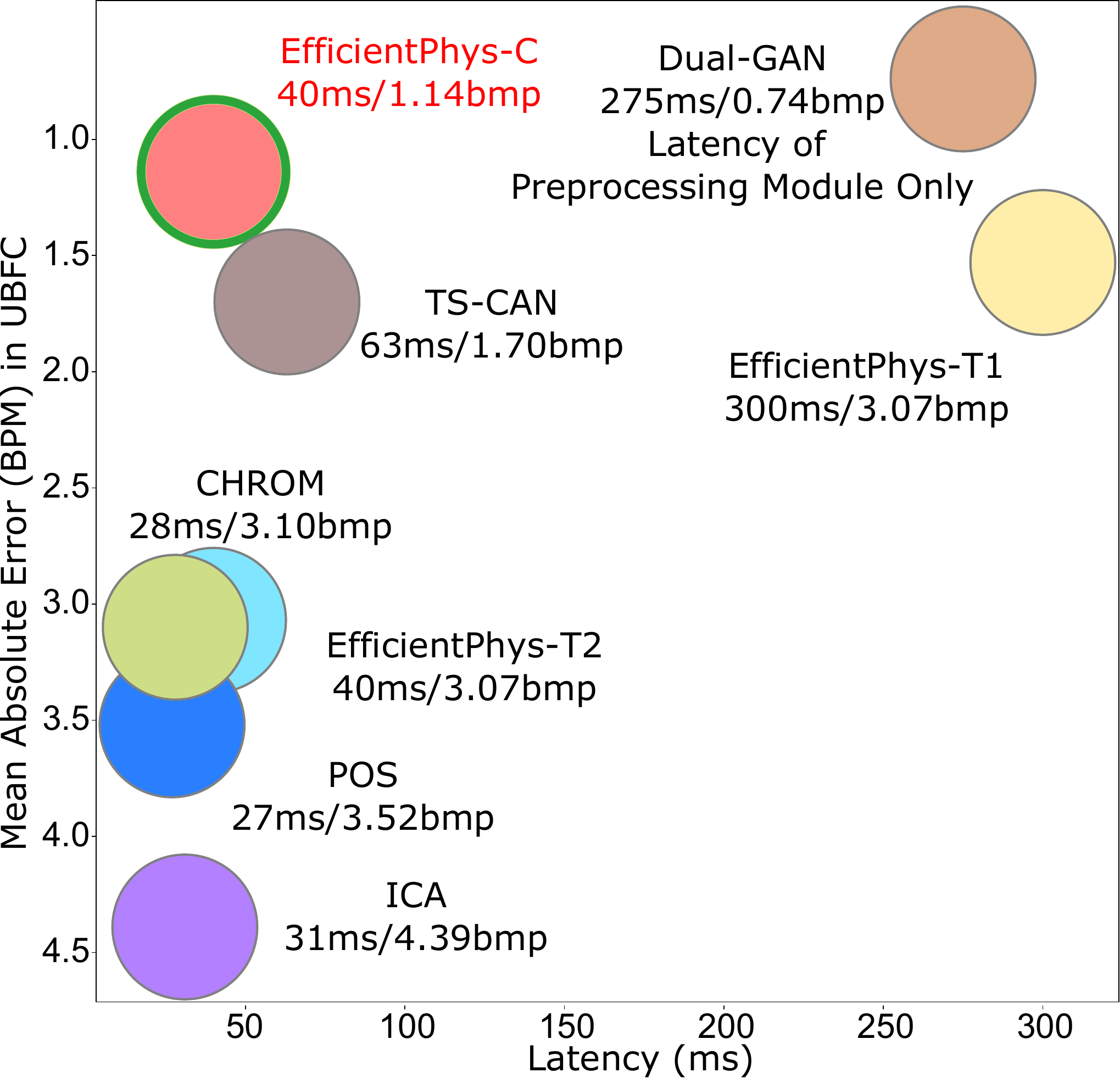}
\caption{Accuracy-Latency Trade-off in eight different methods. Y-axis denotes the MAE error, and X-axis denotes the latency. The methods in the left-top corner have the best accuracy-latency Trade-off.}
\label{fig:latency_acc}
\end{figure}

\begin{table}[t!]
    \small
	\caption{On-Device data preprocessing latency and model inference latency per frame (ms).}
	\vspace{0.3cm}
	\label{tab:latency}
	\centering
	\small
	\setlength\tabcolsep{3pt} 
	\begin{tabular}{r|c|c|c}
	\toprule
	    &\multicolumn{1}{c}{\textbf{Preprocessing}} & 
	    \multicolumn{1}{c}{\textbf{Model}} & 
	    \multicolumn{1}{c}{\textbf{Total}} \\
        \textbf{Method} & (ms) $\downarrow$ & (ms) $\downarrow$ & (ms) $\downarrow$ \\ \hline  \hline 
        EfficientPhys-C & 0 & 40 & 40 \\
        EfficientPhys-T1 & 0 & 300 & 300   \\
        EfficientPhys-T2 & 0 & 40 & 40  \\ \hdashline
        TS-CAN\cite{liu2020multi}& 3 & 60 & 63   \\
        Dual-GAN\cite{lu2021dual} & \textbf{275} & N/A & \textbf{$>275$}  \\
        POS\cite{wang_algorithmic_2017} & 0 & 27 & 27  \\
        CHROM\cite{de2013robust} & 0 & 28 & 28  \\
        ICA\cite{poh2010advancements}& 0 & 31 & 31   \\
       \bottomrule 
       \end{tabular}
       \\
       \tiny
      ms = Preprocessing and model latency on Raspberry Pi 4B per frame.
    \vspace{-0.1cm}
\end{table}

\textbf{EfficientPhys vs. State-of-the-Art.}
In Table \ref{tab:cross_eval_crop} and Table \ref{tab:cross_eval_mmse}, we present results from our proposed EfficientPhys models and the current state-of-the-art neural and signal processing methods. The learning models are all trained on the same datasets (AFRL + Synthetic) and tested on three dataset (UBFC, PURE and MMSE) to test if the model can generalize to videos with a different facial appearance, background, and lighting. To investigate how the depth of the network impacts the Transformer architecture, we created two version of Transformer-based EfficientPhys: T1 and T2. T1 uses the same depth as the Swin Transformer reported in~\cite{liu_swin_2021} ([2, 2, 6, 2]). Each number indicates the number of Swin Transformer blocks as illustrated in Fig. \ref{fig:high_level_comp}. T2 is a more lightweight architecture to enable real-time on-device inference and has a depth of [2, 1]. EfficientPhys-C denotes the Convolution-based EfficientPhys as shown in the Fig.\ref{fig:high_level_comp}. For UBFC and PURE, as Table \ref{tab:cross_eval_crop} illustrates, EfficientPhys-C and EfficientPhys-T1 outperform all the existing methods. As Table \ref{tab:cross_eval_mmse} demonstrates, all the neural methods outperform the signal processing methods. EfficientPhys-T1 and TS-CAN achieved slightly better results than EfficientPhys-C and EfficientPhys-T2. Unfortunately, due to the lack of open source implementation or released models (e.g., Dual-GAN \cite{lu2021dual}), we could not successfully replicate their complicated model architecture and conduct cross-dataset evaluation on this comparison.

To conduct a fair comparison with the current state-of-the-art methods, we followed Dual-GAN \cite{lu2021dual} to train our models only on PURE and to test on UBFC as Table \ref{tab:cross_pure_ubfc} shows. Although Dual-GAN outperforms all of other methods, we argue that the margin is relatively small as both Dual-GAN and EfficientPhys-C achieve a Pearson correlation of 0.99. Moreover, according to American National Standards Institute (ANSI) and Consumer Technology Association's standard \cite{consumer2018physical}, MAPE of $\pm5$ is an acceptable error rate. Various studies have also used this standard to validate FDA approved sensors and systems \cite{nelson2019accuracy,shcherbina2017accuracy,reece2021assessing,carrier2020validity}. All the methods in Table \ref{tab:cross_pure_ubfc} have met this recommended bar.

\begin{table}[t!]
    \small
	\caption{Ablation study on EfficientPhys-C (Top) an EfficientPhys-T1 (Down). Models are trained only on PURE and tested on UBFC.}
	\vspace{0.3cm}
	\label{tab:ablation_study}
	\centering
	\small
	\setlength\tabcolsep{3pt} 
	\begin{tabular}{ccc|r}
	\toprule
        Self-Attention & Diff & BatchNorm &  MAE\\ \hline  \hline 
        \cmark & \cmark & \cmark & 2.13   \\
        \xmark & \cmark & \cmark & 2.43   \\
        \cmark & \cmark & \xmark & 16.06   \\
        \cmark & \xmark &  \xmark &  16.06  \\
       \bottomrule 
       \end{tabular}
    \quad
	\begin{tabular}{cc|r}
	\toprule
        TSM & Normal. Module &  MAE\\ \hline  \hline 
        \cmark & \cmark  & 3.83   \\
        \cmark & \xmark  & 16.10   \\
        \xmark & \cmark  & 11.52   \\
       \bottomrule 
     \end{tabular}
     \\
\end{table}

\textbf{Computational Cost and On-Device Latency.}
Fig. \ref{fig:latency_acc} and the Table \ref{tab:latency} summarize the computational cost of the existing neural methods. Again, due to the lack of open source implementation and complex algorithm design, we were not able to replicate every architecture to benchmark its on-device latency. The results show that EfficientPhys-C only takes 40ms to process a single frame and it does not take any extra computational time to perform preprocessing. On the other hand, due to the complex model architecture and additional time for calculating hand-crafted normalized raw and difference frames, TS-CAN takes 63ms per frame. As mentioned earlier, Dual-GAN has a complicated preprocessing procedure for facial landmark detection, segmentation, color transformation and augmentation. We implemented this and benchmarked the preprocessing module on our platform, and it took 275ms per frame, which is already 7x than the entire computational time of EfficientPhys-C. The estimation network in Dual-GAN also includes 12 2D convolution operations and numerous 1D convolution operations. Thus, we believe it would add a significant amount of computational time on top of the 275ms preprocessing time per frame. The default Transformer-based EfficientPhys (T1) has an unfavorable inference time due to its deep architecture design and takes 300ms to process every single frame. After reducing the depth to EfficientPhys-T2, it can achieve the same inference time as the EfficientPhys-C. However, EfficientPhys-T2 has the poorest performance on all three benchmark datasets.

\textbf{Convolution vs. Transformer in Camera-Based Vitals Measurement.}
Although visual transformers have begun to achieve state-of-the-art performance in some vision tasks, it is not the case for the task of video-based vitals measurement. Based on the results shown in Table \ref{tab:cross_eval_crop} and Table \ref{tab:cross_eval_mmse}, Efficient-C outperforms both Efficient-T1 by 45\% of MAE in UBFC and similar performance in MMSE and PURE, while Efficient-C is more than 7x faster in terms of latency. When we shrink the Transformer-based EfficientPhys to a similar complexity as Convolution-based EfficientPhys, the performance is significantly diminished. The errors from the lightweight Transformer-based EfficientPhys-T2 increased 48\% of MAE in UBFC, 141\% of MAE in PURE and 15\% of MAE in MMSE. These results indicate a shallow transformer architecture struggles to model subtle changes of skin pixels in the video. These finding suggest two potential insights. First, further optimizations will be necessary for transformers to outperform, even relatively shallow, convolutional models in this domain, this is possibly especially true when there is not a large amount of high-quality data available. As previous studies have shown \cite{dosovitskiy2020image}, Transformers usually require more pre-training samples to obtain state-of-the-art accuracy. Unfortunately, currently the amount of data in the field of camera-based vital measurement is limited compared to other visual tasks. Our experiments in Table \ref{tab:cross_pure_ubfc} also support this hypothesis where EfficientPhys-C surpasses both EfficientPhys-T1 and T2 with training only on PURE. We believe synthetic data is one way to help address this issue. Second, the good accuracy-efficiency trade-off for visual transformer might not be scaled to on-device architectures without further work. Since many on-device neural networks require significantly less amount of computing resources to perform real-time operations, scaling the Transformer architecture down is not ideal as our experimental results of EfficientPhys-T2 have shown.

\textbf{Ablation Study. }
We provide ablation studies on various parameters in EfficientPhys-C and EfficientPhys-T1 in Table \ref{tab:ablation_study}. Without the self-attention module, MAE of EfficientPhys-C is increased by 14\%. Without the Normalization Module, in both EfficientPhys-C and EfficientPhys-T1, the MAEs increased by 753\% and 420\%. As Figure \ref{fig:bachnorm} illustrates, the output of the difference layer contains almost black pixels and these results indicate that neural methods are sensitive to the magnitude of pixel values and whether they are zero centered. Finally, without the tensor shift module (TSM) in transformer-based models, the error increased by 300\% which indicates TSM plays an important role in exchanging temporal information and dynamics.

\textbf{Simplifying Last-Mile ML Deployment. }
Numerous real-world applications are driven by novel machine learning algorithms. However, deploying these algorithms on different computing platforms has been extremely challenging for various reasons. One of these is that researchers sometimes only pay attention to the accuracy of the model and ignore the complexity of the last-mile engineering efforts. In this paper, we address this important issue through our one-stop architecture that takes the unprocessed raw frames and directly outputs the desired signal. This elegant and simple design will not only reduce the burden of engineering required for cross-platform implementations, but also will help the research community to replicate and reproduce results.

\textbf{Extensible to Other Signal.} Finally, as another potential upside of our end-to-end design and the low latency, we envision EfficientPhys being applied to various other video-based applications. Since the input of our model is raw frames, we believe EfficientPhys can be easily extended to other tasks such as video-based blood pressure measurement and video understanding \& recognition etc. On the other hand, most of the baseline methods we compared with (e.g., Dual-GAN, PulseGAN) require many custom preprocessing operations for video-based measurement which are less useful in other applications.

\section{Broader Impacts and Ethics Statement}

During the development of EfficientPhys, we wanted to ensure the innovations would not create larger disparities between different populations. By achieving state-of-the-art accuracy and efficiency as well as our simple and elegant design, we believe EfficientPhys will help make camera-based vitals measurement more widely available to the medical research community and broader community in computing. We also believe that this technology can have an especially strong impact in low-resource settings where there are greater barriers to accessing healthcare. We envision our proposed method could, with the appropriate clinical validation and regulatory approval, eventually be used in healthcare applications (e.g., real-time vitals measurement in telehealth appointments). During the COVID-19 pandemic the need for such technology has been clearly highlighted. We contextualize our contributions within the scope of democratizing technology for social good and helping to reduce health disparities with advanced AI technology. However, we are aware that machine learning systems are biased and can propagate inequalities. Before technology such as that presented in this paper is ready for deployment we need to make sure that that is not the case.

\section{Conclusion}

In this paper, we present a novel method called EfficientPhys to enable simple, fast, accurate camera-based contactless vitals measurement. We achieved strong performance with using significant less computational power. With the simple and elegant one-stop design, EfficientPhys also helps address the issue of last-time machine learning deployment and reduces health disparity.

{\small
\bibliographystyle{ieee_fullname}
\bibliography{egbib}
}

\end{document}


\newcommand{\xmark}{\ding{55}}%
\newcommand{\cmark}{\ding{51}}%

\title{EfficientPhys: Enabling Simple, Fast and Accurate Camera-Based Cardiac Measurement} 

\author{Xin Liu\textsuperscript{1}, Brian Hill\textsuperscript{2}, Ziheng Jiang\textsuperscript{1}, Shwetak Patel\textsuperscript{1}, Daniel McDuff\textsuperscript{3}\\
Paul G. Allen School of Computer Science \& Engineering, University of Washington, Seattle, WA, USA\textsuperscript{1} \\
Department of Computer Science, University of California, Los Angeles, CA, USA\textsuperscript{2} \\
Microsoft Research, Redmond, WA, USA\textsuperscript{3} \\ 
{\tt\small \{xliu0, shwetak, ziheng\}@cs.washington.edu} \\
{\tt\small brian.l.hill@cs.ucla.edu, damcduff@microsoft.com}
}

\maketitle
\thispagestyle{empty}

\section{Additional Results}

We further evaluated our proposed models on the MAHNOB-HCI dataset \cite{soleymani2011multimodal} and compared against other state-of-the-art on-device methods in Table \ref{tab:mahnob}. The MAHNOB-HCI dataset contains 527 videos in total with 27 subjects (12 males and 15 females). The ground truth heart rate were computed on the provided ECG waveform, and the sampling rate is 61hz. To calculate the  heart rate from estimated facial PPG, we applied a band-pass filter to the signal with a cutoff frequency of 0.75 and 2Hz (45 beats/minute to 120 beats/minute) and then used FFT to calculate corresponding heart rates. As Table \ref{tab:mahnob} illustrates, EfficientPhys-C achives the best performance across six different on-device methods. Unfortunately, the results of POS are not available on previous literature.

\begin{table}[h!]
    \small
	\caption{Cross-dataset heart rate evaluation on MAHNOB-HCI (beats per minute).}
	\vspace{0.3cm}
	\label{tab:mahnob}
	\centering
	\small
	\setlength\tabcolsep{3pt} 
	\begin{tabular}{r|cccc}
	\toprule
	    & \multicolumn{4}{c}{\textbf{MAHNOB-HCI}~\cite{soleymani2011multimodal}} \\
        \textbf{Method} & MAE$\downarrow$ & MAPE$\downarrow$ &  RMSE$\downarrow$ & $\rho$ $\uparrow$ \\\hline  \hline 
        EfficientPhys-C & 6.16 & 8.39\% & 8.71 & 0.69  \\
        EfficientPhys-T1 & 11.67 & 16.25\% & 14.89 & 0.01  \\
        EfficientPhys-T2 & 11.34 & 15.91\% & 14.15 & 0.09 \\ \hdashline
        TS-CAN\cite{liu2020multi} & 7.47 & 10.13\% & 10.75 & 0.50  \\
        POS\cite{wang_algorithmic_2017} & NA & NA & NA & NA \\
        CHROM\cite{de2013robust} & 13.49 & NA & 22.36 & 0.21 \\
        ICA\cite{poh2010advancements} & NA & NA & 13.60 & 0.36 \\
       \bottomrule 
       \end{tabular}
       \\
       \tiny
      MAE = Mean Absolute Error in HR estimation, MAPE = Mean Absolute Error Percentage in HR estimation, RMSE = Root Mean Square Error in HR estimation, $\rho$ = Pearson Correlation in HR estimation.
    \vspace{-0.1cm}
\end{table}

{\small
\bibliographystyle{ieee_fullname}
\bibliography{egbib}
}